\documentclass[runningheads,a4paper]{llncs}

\usepackage{ifxetex}
\ifxetex
  \usepackage{fontspec}
  \usepackage{polyglossia}
  \setmainlanguage{english}
\else
  \usepackage[utf8]{inputenc}
  \usepackage[T1]{fontenc}
  \usepackage[english]{babel}
\fi

\usepackage{amssymb}
\usepackage{graphicx}
\usepackage{url}
\usepackage{float}
\usepackage{amsmath}
\usepackage{graphicx}
\usepackage{capt-of}
\usepackage{wrapfig}
\usepackage{booktabs}
\usepackage{multirow}
\usepackage{lipsum}
\usepackage{csquotes}
\usepackage{fancyhdr}
\usepackage{lastpage}
\usepackage[all]{nowidow}
\usepackage[inline]{enumitem}
\usepackage[usenames,dvipsnames]{xcolor}

\makeatletter
\def\Hline{%
\noalign{\ifnum0=`}\fi\hrule \@height 1pt \futurelet
\reserved@a\@xhline}
\makeatother

\newcounter{app}
\newcommand{\appendixsection}[1]{%
  \stepcounter{app}%
  \section*{Appendix \theapp: #1}%
  \addcontentsline{toc}{section}{Appendix \theapp: #1}%
}

\begin{document}


%
%
\title{Hibikino-Musashi@Home\\2026 Team Description Paper}

\author{
    Soma Fumoto
    \and Ryohei Kobayashi
    \and Taiki Watanabe
    \and Yuga Yano
    \and Akinobu Mizutani
    \and Naoki Yamaguchi
    \and Ryo Terashima
    \and Ryusei Sakakibara
    \and Kazuma Futami
    \and Keita Okumura
    \and Kosei Isomoto
    \and Koshun Arimura
    \and Yuta Ohno
    \and Tomoaki Fujino
    \and Sojiro Shimada
    \and Yuichiro Tanaka
    \and Takeshi Nishida
    \and Hakaru Tamukoh
}
\authorrunning{Soma Fumoto et al.}
\institute{
Kyushu Institute of Technology\\
The University of Kitakyushu\\
\email{hma@brain.kyutech.ac.jp} \\
\url{https://www.brain.kyutech.ac.jp/~hma/}
}
\maketitle

\begin{abstract} 
This paper presents an overview of the integrated technologies developed for the Open Platform League at RoboCup@Home 2026 in Incheon.
A vision system integrated with learning-based, training-free models enables robust object recognition for both known and unknown objects.
We introduce a RoomPlan-based method that semi-automatically registers furniture geometry and labels to generate semantic maps. 
Additionally, a large language model-based planner interprets the instructions recognized by Whisper and generates primitive skill sequences for execution.
A reservoir-computing framework serves as a lightweight temporal-processing model for action recognition. Its field-programmable gate array-based implementation enables low-power, real-time operation.
This is key to removing operational time constraints for service robots.
Open-source releases, including task environments, dataset tools, and navigation frameworks, enhance reproducibility, reuse, and continuous progress within the RoboCup@Home community.

\end{abstract}


\section{Introduction} 

Hibikino-Musashi@Home (HMA) is a robot development team composed of students from the Kyushu Institute of Technology and University of Kitakyushu in Japan. 
The team was founded in 2010 and has consistently participated in the RoboCup@Home and RoboCup JapanOpen @Home Leagues. 
In addition to RoboCup, the team participated in the World Robot Challenge 2018 and 2020 in the service robotics category of the Partner Robot Challenge (real space), as well as the World Robot Summit 2025 Future Convenience Store Challenge.

HMA addresses various challenges in realising service robots capable of working in real-world environments. 
First, in environments containing diverse objects, robots must adapt to a wide variety of items and appearances.
We use two methods for object recognition, You Only Look Once 11 (YOLO11) trained by our custom dataset, and training-free models such as Grounding-DINO, NanoSAM, and Cutie.
Second, in household environments, service robots must understand the surrounding 3D space to perform tasks appropriately. 
We developed an automated system using a smartphone for constructing semantic maps.
Third, the tasks a robot must perform in household environments change with the situation and the user’s instructions.
We developed a dynamic task planning system based on a large language model (LLM).
Finally, high power consumption is a bottleneck for long-term operation required for practical service robots. 
We introduced a system with reservoir computing (RC) implemented on a field-programmable gate array (FPGA) to achieve low power consumption.

By integrating these components, our goal is to build a service robot system that can robustly handle various environments and tasks.

%
%

\section{Perception}

\subsection{Object Recognition}
\label{ss:object-detection}
Object recognition is a fundamental component of the robot system. To achieve robustness and adaptability, we employed a combination of learning-based and training-free methods. 
The learning-based model YOLO11~\cite{yolo11_ultralytics} provides robust recognition of known objects.
In contrast, training-free models, such as Grounding DINO~\cite{G_DINO} and NanoSAM~\cite{nano_sam}, enable prompt-based segmentation of unknown objects.
In addition, the training-free method Cutie~\cite{cutie2024} supports object detection and tracking using predefined masks, enabling robust perception through prior mask preparation.
These complementary methods enable flexible and reliable perception across diverse environments.

\subsubsection{YOLO} 
We adopted YOLO11 \cite{yolo11_ultralytics} as our primary detection model.
High-quality training data is essential for maximizing YOLO’s performance, so we constructed a large-scale dataset using a PyBullet-based 3D simulation system \cite{coumans2016pybullet,ono2022ar}.
Each task object was scanned with a LiDAR-equipped smartphone, reconstructed as a 3D mesh, and imported into the simulator.
We rendered diverse synthetic images by randomizing camera poses, lighting, background textures, and furniture layouts, ensuring sufficient variation for robust training.
After building the dataset, we fine-tuned YOLO11 on the synthesized images.
To enable real-time processing, we optimized the fine-tuned YOLO model as a TensorRT engine with INT8 quantization, significantly reducing inference latency and memory usage while preserving detection accuracy.

\subsubsection{Combination of Grounding DINO and NanoSAM} 
As a training-free method, we integrated Grounding DINO and NanoSAM to construct a text-prompted two-stage object recognition pipeline. 
Grounding DINO enables the detection of diverse objects by simply specifying text prompts, thereby eliminating the need for large task-specific datasets. 
This model represents semantic and appearance attributes, such as color, material, and category, as text prompts and determines the location of the target object based on this input. 
Grounding DINO produces bounding boxes for a given image, and then NanoSAM outputs object masks for grasp point estimation in manipulation.


\subsubsection{Cutie} 

\begin{figure}[tb]
\begin{center}
    \includegraphics[width=0.8\columnwidth]{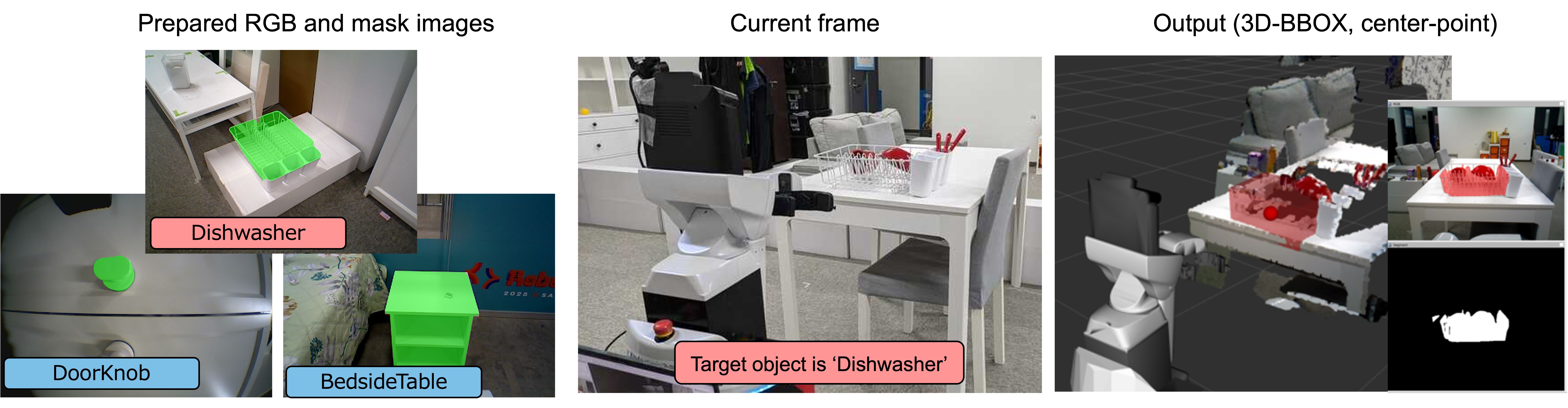}
    \caption{Overview of Cutie.}
    \label{fig:cutie_overview}
  \end{center}
  \end{figure}

We utilized Cutie to detect quasi-static environmental objects, such as desks and shelves. 
Cutie is a tracking method that segments a target object within images using a reference RGB image and a segmented binary mask image of the object.
We developed a framework that prepares the necessary RGB images and masks for this tracking method in advance and calls them as needed for tracking purposes.
Figure~\ref{fig:cutie_overview} shows an overview of the Cutie-based object detection framework.
As a preliminary preparation, we captured RGB images corresponding to each object and created mask images for the target objects. 
We stored the captured RGB images and masks in a database in a callable format under specific names.
During inference, the framework retrieves the corresponding image pair of the target object and segments the object in the image obtained from the robot.
Based on the obtained segmented image, the framework outputs the 3D shape and position of the object. 
All implementations are publicly available on GitHub and can be easily integrated with real robots and simulator environments~\footnote{\url{https://github.com/Hibikino-Musashi-Home/cutie-ros}}.



\subsection{Semantic Map} 
\label{subsection:semantic_map}
\begin{figure}[tb]
  \begin{center}
    \includegraphics[width= 0.6\columnwidth]{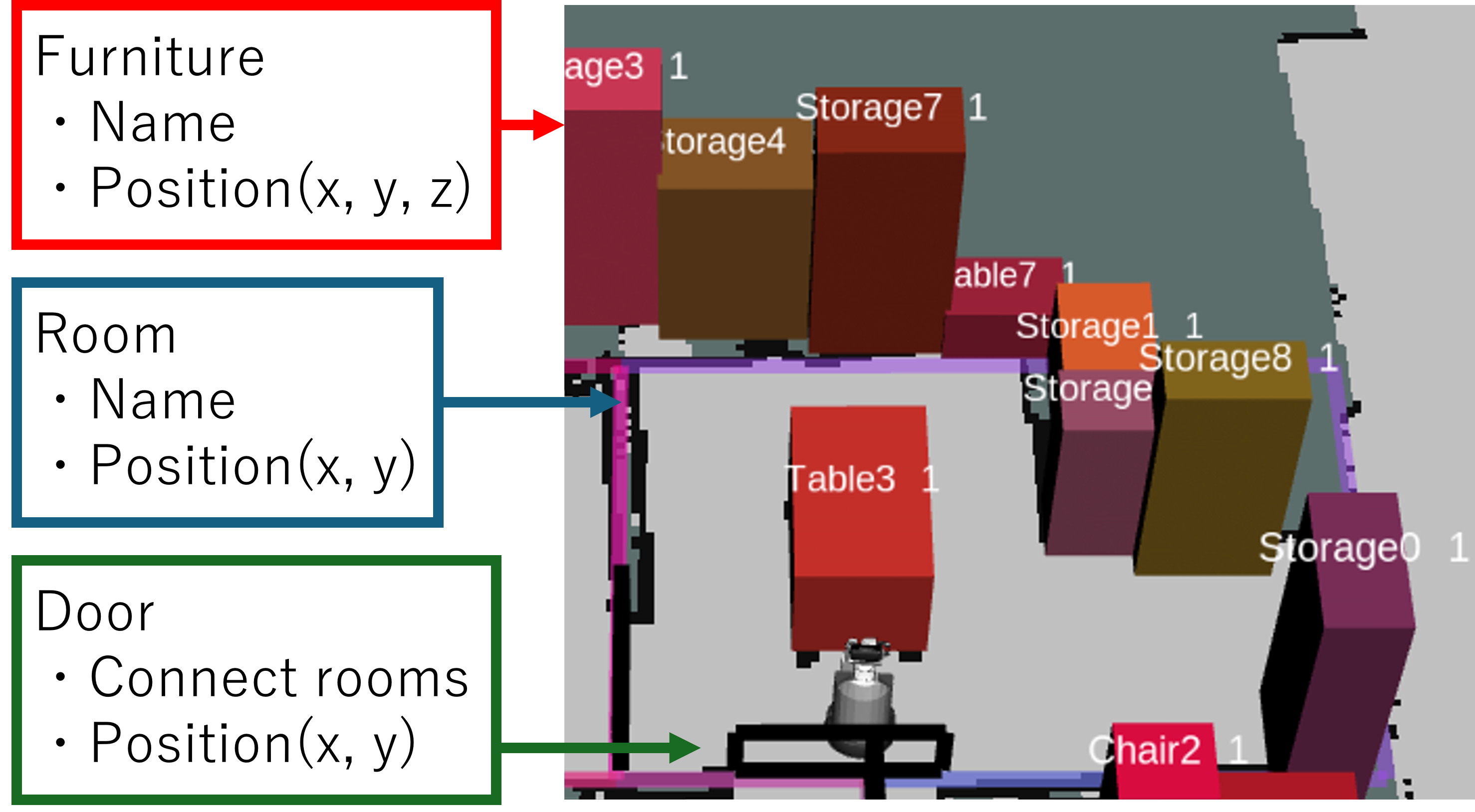}
    \caption{Example of a semantic map.}
    \label{fig:semantic-map}
  \end{center}
\end{figure}

A semantic map is a technique that assigns semantic labels—such as doors, sofas, or televisions—to each location within the geometric space created by a robot during mapping.
Figure \ref{fig:semantic-map} shows our semantic map, which adds information about rooms, furniture, and doors to the geometric map.
Each room and piece of furniture is assigned a label containing its name and positional information, represented by an array of 3D coordinates denoting the vertices of the outline.
The robot utilizes semantic information to determine its own position or that of a human.
To determine the room in which the robot is currently located, polygonal area detection is performed. This detection identifies a room by referencing the outline information of all rooms in the environment.

\subsubsection{RoomPlan} 
Such semantic maps must be recreated when the environment changes, such as when furniture is moved. Manually creating and updating such maps takes significant time and effort, and human errors, such as typos, can cause unexpected behavior of the robot.
To address this issue, we use RoomPlan~\cite{AppleRoomPlan}, an iPhone/iPad application that uses the device camera and LiDAR scanner to automatically create a 3D floor plan with room dimensions and furniture information.
We semi-automated the creation of the ``Furniture'' part (Fig.~\ref{fig:semantic-map}) in two steps:
\begin{enumerate}[label=(\arabic*)]
  \item We scan the room using RoomPlan on a LiDAR-equipped iPhone/iPad to generate a Universal Scene Description (USD)~\cite{OpenUSD}, in which furniture is approximated by cuboids.
  \item We then analyze the obtained USD to extract the name, position, dimensions, and pose of each furniture item and automatically register them into the ``Furniture'' element.
\end{enumerate}
The extracted furniture information is overlaid onto the environmental map, and the furniture areas are colored for visualization.
This method reduces errors originating from manual input and allows the information necessary for path planning to be prepared quickly.

\subsection{LLM-Based Task Planning from Natural Language Instructions} 
\begin{figure}[tb]
  \begin{center}
  \includegraphics[width=0.9\columnwidth]{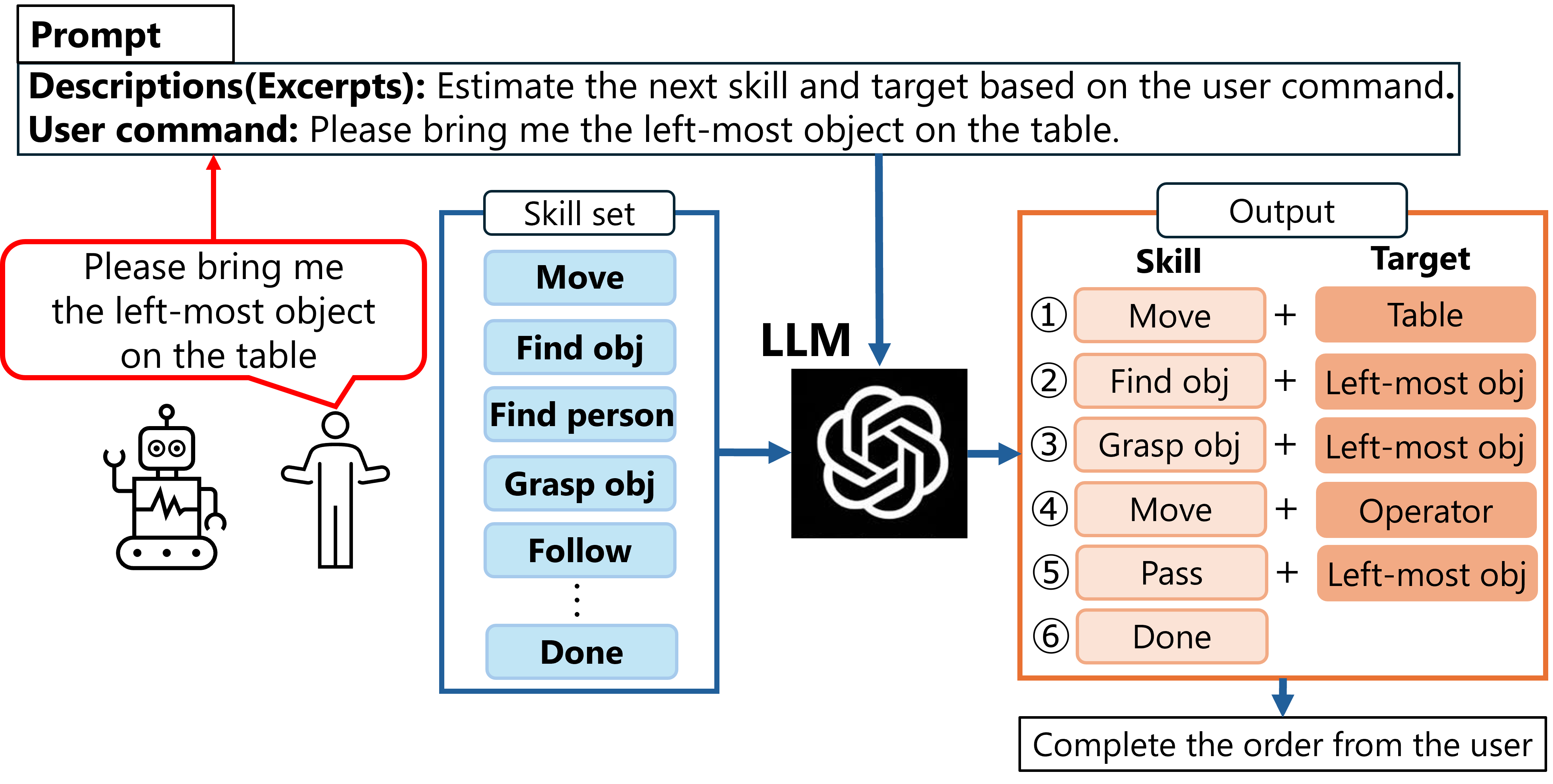} 
  \caption{Overview of the task planning system.}
  \label{fig:task_planning}
  \end{center}
\end{figure}

In real-world environments, robots must understand human instructions and intentions to perform autonomous actions.
We propose a task planning system that utilizes an LLM.
Figure \ref{fig:task_planning} illustrates an overview of the proposed task planning system.
First, the user’s voice command is converted to text using the speech recognition model Whisper \cite{radford2022robust}. Subsequently, the LLM \cite{gpt-4o} is provided with the robot role description, environmental information, user command, and available skill information.
Skill information refers to the basic actions that a robot can take to perform tasks.
We provide 12 predefined skills that the robot selects and combines to achieve the task via the LLM.
The representative skills include \texttt{find\_obj}, \texttt{grasp}, \texttt{move}, and \texttt{observe\_obj}.
\texttt{find\_obj} is the object detection skill described in Section \ref{ss:object-detection}, while \texttt{grasp} is the grasp point estimation skill based on 3D point cloud information.
Consider the command ``Bring me the cup on the desk.'' as an example.
The LLM sequentially combines and selects \texttt{move}, \texttt{find\_obj}, and \texttt{grasp}, planning the movement, detection, and grasping of the target object.
The proposed task planning system enables robots to autonomously perform tasks, from understanding the environment to planning actions and executing movements.

\section{Action Recognition by Reservoir Computing}
\label{ss:reservoir-computing} 
\subsubsection{Reservoir Computing}
RC is a framework designed for time-series processing.
Typical RC models include echo state networks (ESNs) \cite{jaeger2001echostate} and liquid state machines \cite{MAASS2004593}.
An RC model consists of an input layer, a reservoir layer, and an output layer.
The RC is a sparsely connected nonlinear dynamical system that projects input signals into a high-dimensional state space, from which the output layer linearly separates relevant patterns. A key property of RC is that both the input–reservoir weights and the recurrent reservoir weights are randomly initialized and kept fixed—even during training.
Training is applied only to these readout weights. Since only the output layer is trained, the overall architecture is simple and hardware-oriented, making it a practical method for resource-constrained environments. 




%


\subsubsection{Action Recognition implemented on an FPGA} 
Service robots must perform multiple recognition methods under limited computational resources, and relying solely on GPUs increases power consumption and reduces overall responsiveness. 
To maintain both performance and energy efficiency under such constraints, we implemented a hand-waving action recognition system with RC on an FPGA accelerator.
FPGAs are reconfigurable semiconductor devices that can be programmed to configure their internal circuit structures, enabling low-power and high-performance processing by optimizing data flow for specific applications. 

\begin{figure}[tb]
\begin{center}
    \includegraphics[width=0.8\columnwidth]{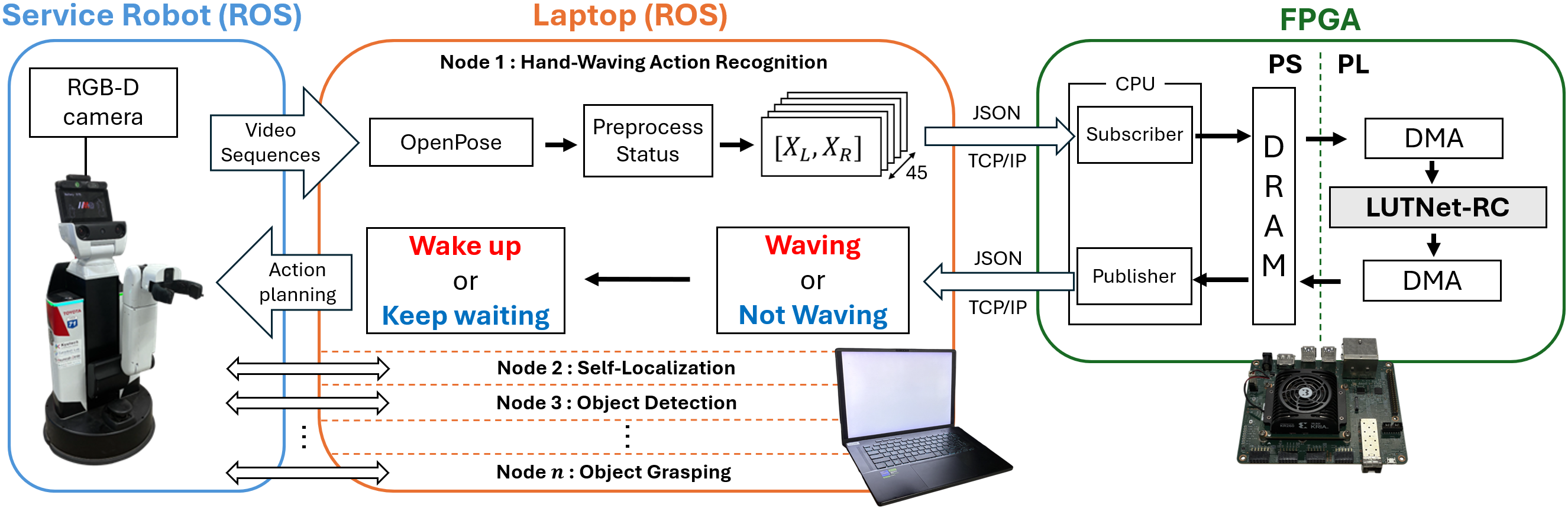}
    \caption{System architecture of the FPGA-accelerated hand-waving action recognition.}
    \label{fig:system_architecture}
\end{center}
\end{figure}

As shown in Fig. \ref{fig:system_architecture}, the proposed system consists of three main components: a service robot and a laptop, both running ROS, and an FPGA-based accelerator.
The service robot is equipped with an RGB-D camera that captures video sequences of the hand-waving actions. 
These images are processed on the laptop to extract skeletal information and convert it into time-series data ($X_L$, $X_R$), representing left and right hand movements. 
The laptop transfers the formatted data to the FPGA via an Ethernet cable using JavaScript Object Notation (JSON) over Transmission Control Protocol/Internet Protocol (TCP/IP). 
The FPGA performs recognition processing and returns the results to the laptop, which sends them back to the service robot.


\section{Reusability} 

We have been actively involved in the HSR community, promoting the development and open-sourcing of robotic systems. 
Since 2024, the HSR community has started the HSRT-X project to develop a foundation model for HSR \cite{matsushima2024hsrtx}. To promote effective data collection, we developed a semi-automatic data collection system based on our previously released TidyUp task system from the World Robot Challenge and RoboCup 2021 online competition \footnote{\url{https://github.com/Hibikino-Musashi-Home/hma_wrs_sim_ws}}. 
Trajectory and image data are collected while the robot tidies up objects automatically using an object detection system, such as YOLO. Instructions in natural language are generated based on the recognition result, and humans check the instruction if the recognition result is incorrect. This system is available in the HSR community as open-source software \footnote{\url{https://git.hsr.io/hsrtx/kyutech-hsrtx-tu.git}}. 

Furthermore, we have been developing and maintaining a practical navigation system that performs robustly across diverse environments by updating an existing ROS~1 implementation by team Pumas \cite{robot_navigation}.
We have integrated several improvements and error-handling mechanisms introduced by our team during RoboCup 2024 \cite{pumas_nav_rk}.
These achievements have been further extended and maintained, in collaboration with co-developers, as a ROS~2 Humble–compatible version \cite{contreras:2025}\cite{pumas_nav2}, which has been open-sourced to enhance reusability and contribute to the broader RoboCup community.

\section{Conclusions} 
This paper introduced the key technologies developed to realize an adaptive and energy-efficient home service robot.
Our system integrates multiple perception modules—including learning-based and training-free object recognition, 3D semantic mapping, and an LLM-based task planner—which together enable robust operation across diverse household environments.  
In addition, we employ RC for action recognition and implement selected components on FPGA hardware to improve overall energy efficiency.
Furthermore, several technologies have been released as open-source software, contributing to the research community and improving their reproducibility.
The team will continue to refine each module through ongoing competition participation, contributing to the advancement of practical and reusable service robot technologies. 

\section*{Acknowledgement} 
This paper is based on results obtained from a project, JPNP16007, and JPNP20004 commissioned by the New Energy and Industrial Technology Development Organization (NEDO).
This paper is also supported by Joint Graduate School Intelligent Car, Robotics \& AI, Kyushu Institute of Technology student project, YASKAWA electric corporation project, JSPS KAKENHI Grant Numbers 23H03468 and 23K18495, JST ALCA-Next Grant Number JPMJAN23F3, and JST SPRING Grant Number JPMJSP2154.


\bibliography{ref}
\bibliographystyle{splncs04} 

\newpage


\appendixsection{Robot's Hardware Description} 

Our service robot, the TOYOTA Human Support Robot (HSR), as shown in Figure~\ref{fig:hsrc_ex}, consists of the following components:

\noindent
\begin{minipage}[t]{0.55\textwidth}  
\vspace{0pt}
\textbf{HARDWARE:}
\begin{itemize}[leftmargin=2em]
    \item \textbf{Mobile base:} Differential-drive base with two wheels and two casters, capable of omnidirectional movement through rotational control.
    \item \textbf{Manipulator:} A single 5-DOF lightweight arm with torque sensors and a gripper equipped with tactile sensors for safe human–robot interaction, including an arm-lift mechanism that enables height adjustment from approximately 0.75 to 1.35~m.
    \item \textbf{Head:} 2-DOF (pan and tilt) mechanism supporting visual and depth sensors.
    \item \textbf{Speaker:} Speaker to generate synthetic speech.
    \item \textbf{HDR Cameras:} TIER IV C2 HDR camera providing high-dynamic-range imaging robust to glare and low-light scenes.
    \item \textbf{RGB-D Cameras:} 
        \begin{itemize}
            \item \textbf{Head:} Orbbec Gemini 336L RGB-D camera for wide-range perception and object recognition.  
            \item \textbf{Hand:} RealSense D401 RGB-D camera for close-range perception and grasping tasks.
        \end{itemize}
    \item \textbf{Microphone and Speakers:} Built-in microphone array and stereo speakers for voice interaction.
    \item \textbf{Laser:} Hokuyo laser rangefinder URG-04LX-UG0.
    \item \textbf{Other Sensors:} Wheel encoders, IMU, and force-torque sensors for localization and safe manipulation.
    \item \textbf{Embedded System:} Onboard Intel i7 processor for real-time control.
\end{itemize}
\end{minipage}
\hfill
\begin{minipage}[t]{0.4\textwidth}  
\vspace{0pt}
\centering
\includegraphics[width=\linewidth]{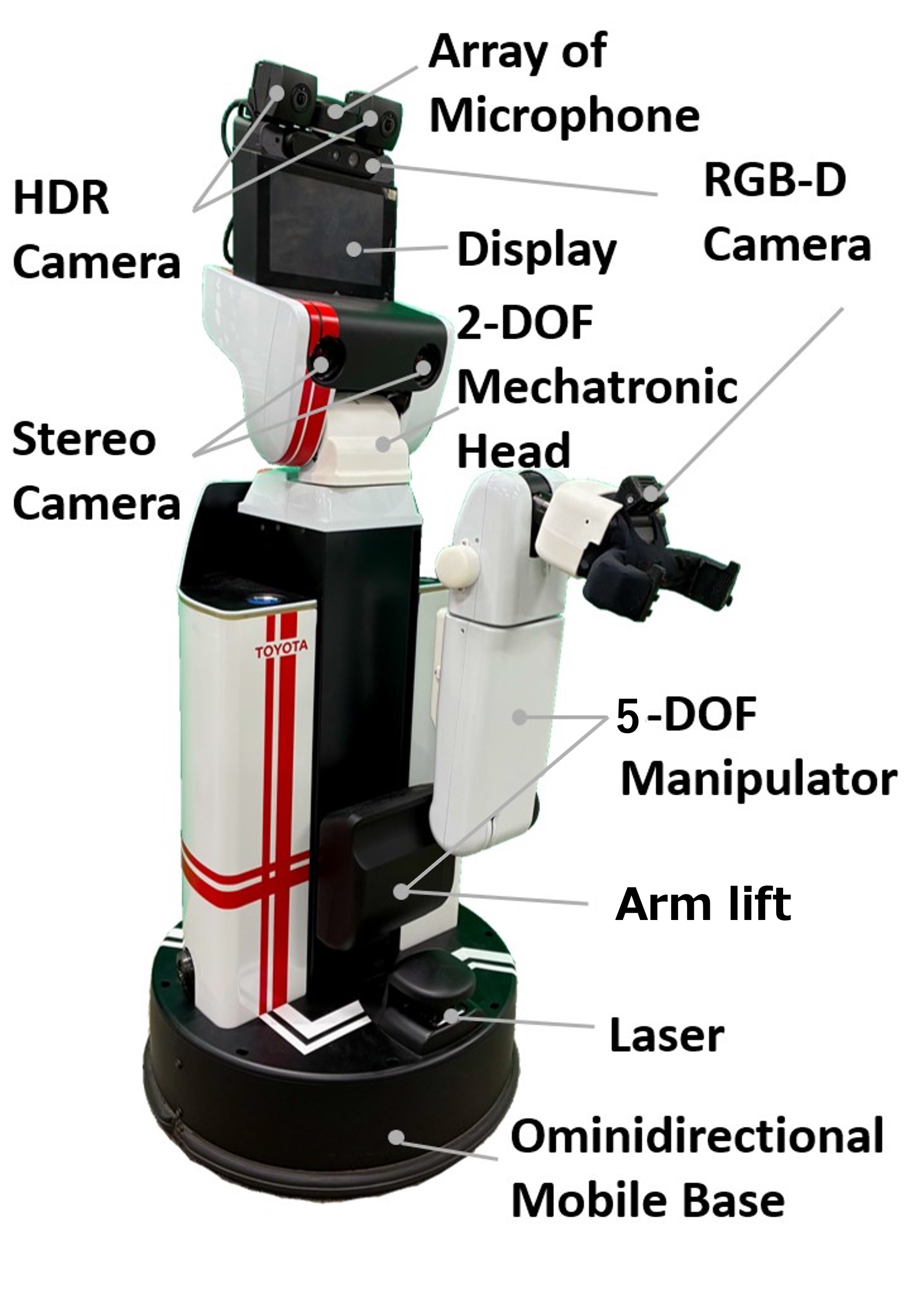}
\captionof{figure}{TOYOTA Human Support Robot (HSR).}
\label{fig:hsrc_ex}
\end{minipage}
\appendixsection{Robot's Software Description} 

The software stack for the robot system is as follows:

\begin{itemize}
	\item Host OS: Ubuntu 22.04
    \item Container: Singularity \cite{singularity}
	\item Middleware: ROS 2 Humble
	\item State management: YASMIN (ROS 2)
    \item Object detection: YOLO11 and NanoSAM \cite{yolo11_ultralytics} \cite{nano_sam}
    \item Object tracking: Cutie \cite{cutie2024}
	\item Speech recognition: Whisper \cite{radford2022robust} and Voice Activity Detection \cite{pyvad2016}
    \item Human detection: OpenPose and StrongSORT \cite{Cao2018OpenPose} \cite{Du_2021_ICCV}
    \item Human action recognition: ESN-based Human action recognition system \cite{jaeger2001echostate} \cite{hand_waving_recognition}
	\item Mapping: Slam Toolbox \cite{slam_toolbox}
    \item Localization: emcl2\_ros2 \cite{emcl2ros2}
    \item Navigation: Pumas Nav2 \cite{contreras:2025}
\end{itemize}

The following are the specifications of the laptop mounted on our HSR:
\begin{itemize}
        \item Model name: ROG Zephyrus M16 GU604VY
        \item CPU: Intel Core i9-13900H
        \item RAM: 32GB 
        \item GPU: NVIDIA GeForce RTX 4090 (16GB) 
\end{itemize}

\appendixsection{Competition results} 
Table \ref{tab:result} lists the results achieved by our team during recent competitions. We have been participating in the RoboCup and World Robot Challenges for several years. Our team has won several prizes and academic awards since then.

\begin{table}[t]
\begin{center}
\caption{Results of recent competitions [DSPL, domestic standard-platform league; JSAI, Japanese Society for Artificial Intelligence; OPL, open-platform league].}
\label{tab:result}
\begin{tabular}{l|l} \hline
	\multicolumn{1}{c|}{Competition} & \multicolumn{1}{c}{Result} \\ \hline \hline
  
        2017 Nagoya & {\bf @Home DSPL 1st} \\
                                   & @Home OPL 5th \\ \hline

        JapanOpen 2018 Ogaki & @Home DSPL 2nd \\
                                   & \textbf{@Home OPL 1st} \\
                                   & JSAI Award \\ \hline
    
        2018 Montreal & \textbf{@Home DSPL 1st} \\
                                   & P\&G Dishwasher Challenge Award \\ \hline


        2019 Sydney & @Home DSPL 3rd \\ \hline

        JapanOpen 2019 Nagaoka &\textbf{@Home DSPL 1st} \\
                                        & \textbf{@Home OPL 1st} \\ \hline
        JapanOpen 2020 & @Home Simulation Technical Challenge 2nd \\
                                        & \textbf{@Home DSPL 1st} \\
                                        & @Home DSPL Technical Challenge 2nd \\
                                        & \textbf{@Home OPL 1st} \\
                                        & \textbf{@Home OPL Technical Challenge 1st} \\
                                        & @Home Simulation DSPL 2nd \\ \hline
        Worldwide 2021 & @Home DSPL 2nd \\
                                        & \textbf{@Home Best Open Challenge Award 1st} \\
                                        & \textbf{@Home Best Test Performance: } \\
                                        & \textbf{    Go, Get It! 1st} \\
                                        & \textbf{@Home Best Go, Get It! 1st} \\ \hline
        Asia-Pacific 2021 Aichi & \textbf{@Home DSPL 1st}\\
                                        & \textbf{@Home OPL 1st} \\ \hline
        JapanOpen 2021 & \textbf{@Home DSPL 1st} \\
                                        & \textbf{@Home DSPL Technical Challenge 1st} \\
                                        & @Home OPL 2nd \\
                                        & \textbf{@Home OPL Technical Challenge 1st} \\ \hline
        2022 Bangkok & @Home DSPL 3rd\\ 
                                        & @Home Best Open Challenge Award\\
                                        & @Home Robo-host \\
                                        & (Party-Host highest score in Stage I tasks)\\ \hline
        JapanOpen 2022 Tokyo & \textbf{@Home DSPL 1st} \\
                                        & @Home DSPL Technical Challenge 2nd \\
                                        & @Home OPL 2nd \\
                                        & \textbf{@Home OPL Technical Challenge 1st} \\ \hline
        JapanOpen 2023 Shiga & \textbf{@Home DSPL 3rd} \\
                                        & \textbf{@Home DSPL Open Challenge 1st} \\
                                        & @Home OPL 2nd \\
                                        & \textbf{@Home OPL Open Challenge 1st} \\ \hline
        2023 Bordeaux & @Home DSPL 2nd\\ \hline
	JapanOpen 2024 Shiga & @Home DSPL 2nd \\
					& @Home DSPL Open Challenge 2nd \\
					& @Home OPL 2nd \\
					& \textbf{@Home OPL Open Challenge 1st} \\
					& @Home Simulation OPL 2nd \\ \hline
	2024 Eindhoven & \textbf{@Home DSPL 1st} \\
                                        & @Home GPSR Overbot\\
                                        & (Best in GPSR \& EGPSR) 1st\\
                                        & @Home Smoothest, Safest Navigation Award 1st\\
                                        & @Home Robo-host\\
                                        & (Party-Host highest score in Stage I tasks) 1st\\
                                        & @Home Robo-butler\\
                                        & (Housekeeper highest score in Stage II tasks) 1st\\ \hline
\end{tabular}
\end{center}
\end{table}

\clearpage

\begin{table}[t]
\begin{center}
\begin{tabular}{l|l} \hline
	\multicolumn{1}{c|}{Competition} & \multicolumn{1}{c}{Result} \\ \hline \hline
        JapanOpen 2025 Shiga & @Home DSPL(Advanced) 2nd \\
					& @Home DSPL Open Challenge 3rd \\
                                        & \textbf{@Home Simulation OPL Open Challenge 1st} \\
					& \textbf{@Home OPL Open Challenge 1st} \\
					& \textbf{@Home Simulation OPL 1st} \\ \hline
	2025 Salvador & \textbf{@Home DSPL 1st} \\
                                        & Exceptional Navigation Award (Smoothest, Safest, Human-friendly) First Place \\ \hline
\end{tabular}
\end{center}
\end{table}
\appendixsection{Links}

\begin{itemize}
  \item Team Video \\ \url{https://youtu.be/C-csIXvmuvA} \\
  \item Team Website \\ \url{https://www.brain.kyutech.ac.jp/~hma} \\
  \item GitHub \\ \url{https://github.com/Hibikino-Musashi-Home} \\
  \item Facebook \\ \url{https://www.facebook.com/HibikinoMusashiAthome} \\
  \item YouTube \\ \url{https://www.youtube.com/@hma_wakamatsu}
\end{itemize}


\end{document}